\documentclass[10pt,a4paper]{article}

\usepackage{inputenc}
\usepackage{graphicx}
\usepackage{amsmath}
\usepackage{caption}
\usepackage{latexsym}
\usepackage{morefloats}
\usepackage{floatrow}
\usepackage{multicol}
\usepackage{multirow}
\usepackage{float}
\floatstyle{plaintop}
\restylefloat{table}
\usepackage{amsthm}
\usepackage{enumitem}
\usepackage{graphicx}
\usepackage{subcaption}
\usepackage{enumitem}
\newlist{steps}{enumerate}{1}
\setlist[steps, 1]{label = Step \arabic*:}
\newtheorem{Example}{Example}
\usepackage{epstopdf}

\begin{document}

\title{\textit{iDCR}: Improved Dempster Combination Rule for Multisensor Fault Diagnosis}
\author{Nimisha~Ghosh,
~Sayantan~Saha,~Rourab~Paul\\
	Institute of Technical Education and Research,\\
	Siksha 'O' Anusandhan (Deemed to be University), Bhubaneswar 751030, Odisha, India
}
\maketitle	
\thispagestyle{empty}
\pagestyle{empty}
\begin{abstract}
Data gathered from multiple sensors can be effectively fused for accurate monitoring of many engineering applications. In the last few years, one of the most sought after applications for multisensor fusion has been fault diagnosis. Dempster Shafer Theory of Evidence along with Dempster's Combination Rule is a very popular method for multisensor fusion which can be successfully applied to fault diagnosis. But if the information obtained from the different sensors shows high conflict, the classical Dempster's Combination Rule may produce counter-intuitive result. To overcome this shortcoming, this paper proposes an improved combination rule for multisensor data fusion. Numerical examples have been put forward to show the effectiveness of the proposed method. Comparative analysis has also been carried out with existing methods to show the superiority of the proposed method in mutisensor fault diagnosis. 
\end{abstract}


\section{Introduction}
\label{intro}
With the growing demands of surveillance systems and the booming market of Internet of Things, sensors have become an imperative part of everyday life. Thus, the reliability on sensors is increasing with each passing day. However, traditional single sensor systems have not really been able to match up to the sensor performance. For instance, if with three sensors it is attempted to detect the fault in a machine, where the first one shows a temperature of $35^0$ C, second one $42^0$ C and the third one $37^0$, the machine is considered as faulty if the temperature is above $40^0$ C. So, it is difficult to judge from the sensor values if the machine can be considered as faulty. If information from only one sensor is taken into account, then there is high probability of information loss and further varied decision results will be acquired. This can be attributed to the conflicting results of the sensors. In reality, the information from different sensors may be inaccurate, fuzzy and conflicting, resulting in a difficult decision making. To mitigate these problem, Bayesian theory~\cite{CAI}, fuzzy sets~\cite{islam}	, Z-numbers~\cite{Jiang}, D-numbers~\cite{LIUChen}, evidence theory etc. have been put forth. In~\cite{CAI}, to build a fault diagnosis model, the authors have used a Bayesian network with two layers namely a fault layer and a fault symptom layer. In~\cite{islam}, fuzzy set theory has been applied for fault diagnosis. To combat data uncertainty,~\cite{Jiang} has used Z-number to combine the data from different sensors. The authors have used D-number theory in~\cite{LIUChen} to handle uncertain information for risk analysis. This work has shown that D-number is effective in combining data from different sensors. But for the error in the sensor data, the information may result in high conflict, thus culminating in error prone fusion result. Dempster-Shafer evidence theory can deal with this conflicting information.

Dempster-Shafer theory of evidence (DSTE)~\cite{Dempster2008,shafer197} is a mathematical concept for combinining data from different sensors. It is a generalisation of Bayesian method and offers better fusion result than most of the aforementioned methods. Thus, evidence theory has been incorporated in many applications to deal with incomplete and uncertain information. An agent-oriented intelligent fault diagnosis system has been proposed in~\cite{LUO2012}. Belief entropy has been used in~\cite{yuan2016} by considering sensor reliability and assigning different weights using a distance function. The authors in~\cite{jiang2016} have used Deng's entropy to measure and modify the evidence by using discounting coefficients. Finally, Dempster's Combination Rule (DCR) has been used to fuse the modified evidence to detect fault in a motor rotor. A novel fault diagnosis method of proton exchange membrane fuel cell (PEMFC) system has been proposed in~\cite{Liu} which combines Extreme Learning Machine (ELM) and Dempster-Shafer theory of evidence. The failure diagnosis model is taken care of by ELM and the diagnostic output is combined by DSTE. Support vector machine (SVM)-evidence theory model has been used in~\cite{HUI2017} to solve information gathered by every SVM model to increase the accuracy of classification. The resulting model has been shown to improve the accuracy by removing all conflicting information from the original SVM model. Thus, it can be easily concluded that evidence theory can be successfully used for fault diagnosis. Keeping this into consideration, this work proposes an effective method to  improve traditional DCR for information fusion.

In this work, information gathered from different sensors are considered as diagnosis evidence and the fault diagnosis model is based on evidence fusion and the corresponding decision. Also, an improved DCR has been put forward to deal with uncertainity in the data expressed by weighted Deng entropy. The rest of the paper is organised as follows: Section~\ref{rs} gives an overview of the relevant literature pertaining to improved DCR. In Section~\ref{prelim}, some concepts of DSTE and weighted Deng entropy are elaborated. Section~\ref{pm} presents the proposed method to deal with conflicting and uncertain information followed by Section~\ref{app} which illustrates an application of fault diagnosis system. Finally, the paper is concluded in Section~\ref{con}.
 
\section{Related Study}
\label{rs}
A lot of methods have come up in the past to deal with the uncertainty and the conflict in the data. The idea is to redistribute the conflict factor $K$ which was attempted by Yager in~\cite{yager}. In this work, the $K$ value has been reallocated to an unknown domain and also a modified DCR has been proposed. However, this method was not very capable to reduce the uncertainty of the fused result. Jiang \textit{et al}.~\cite{jiang2018} modified~\cite{yager} to fuse the basic probability assignment (BPA). Sun \textit{et al}.~\cite{Sun} used valid coefficients to consider partial conflicts of the evidence. Li \textit{et al}.~\cite{li2001efficient} considered equality of each of group of evidence to distribute the conflict factor $K$ and used weighted average support degree to achieve the same. Frame of Discernment (FD) (explained in Section~\ref{prelim}) was extended in~\cite{Smarandache2005} to mitigate the conflicting behaviour of information. Martin \textit{et al}.~\cite{Martin} used a combination rule to propose a discounting procedure to deal with partial conflicts. Reliability of evidence was calculated in~\cite{LiangX} to modify the combination model. Li and Gou~\cite{WLLi} extended~\cite{Sun} and~\cite{li2001efficient} by taking into consideration the conflict distribution of the proposition. 

There have been other works as well which did not modify the traditional DCR. Rather, they worked with preprocessing of original evidence. To deal with conflicting evidence, Murphy~\cite{Murphy} used average of BPA, but failed to consider the relationship between the evidences. In~\cite{JOUSSELME}, the authors proposed a new method of distance measurement (Jousselme's distance) which proved to be an effective method for finding the correlation among the evidence. But, if all the elements of FDs are a part of Focal Elements (FE) (explained in Section~\ref{prelim}), the calculation will get a bit tricky due to the rise in cardinality. Pignistic probability distance was used by~\cite{ChenY} to preprocess the original BPA and calculate the weighted evidence. Similarity measurements were used by Zhang \textit{et al}.~\cite{ZhangYing} to calculate support degree and weighted factor of each evidence. Zhang \textit{et al}.~\cite{Zhang2014} further used $n$-dimensional Pignistic probability vector based on cosine theorem and the weight factor was determined by considering the angle measurements between the evidences. Classification algorithms for ensemble learning along with evidence theory was used by~\cite{TABASSIAN2012} to deal with the uncertainty in the information. BPA based on normal distribution was put forth by Xu \textit{et al}. in~\cite{XU2013}. In this work, training data was used to construct the normal distribution and then a BPA function was determined by considering the relation between the normal distribution model and the  test data. Zhang \textit{et al}.~\cite{ZHANGChen} determined BPA from the distance between the core data samples and the selected data. Lu \textit{et al}.~\cite{RLu} used Mahalanobis distance to propose a combination rule which worked well for highly conflicting evidences. Supporting probability distance was used by~\cite{YuCholo} to work with the combination rule. In~\cite{YLin} Lin \textit{et al}. used Euclidean distance measurement to distinguish between different evidences. Wang \textit{et al}.~\cite{ZWang} have used belief entropy to reduce the uncertainty on the evidences, Though, many methods have been applied to mitigate the high conflict results, scope for improvement still persists as has been shown in this work.
\section{Preliminaries}
\label{prelim}
Before delving deep into the methodology, some preliminaries are outlined:
\subsection{Dempster-Shafer Evidence Theory}
Uncertain reasoning method, popularly known as Evidence Theory, has found its applications in a lot of areas like fault diagnosis~\cite{YLin,ZWang}, uncertainty modelling, decision making etc. Dempster-Shafer Evidence Theory, also referred to as theory of belief functions, has been greatly used in the field of uncertainty modelling. The basic concepts of DS theory are introduced next.

\textbf{Frame of discernment} is the complete set of all hypotheses where the elements are mutually exclusive and the set is exhaustive. It is represented by:
\begin{equation}
\chi = \{F_1,F_2,\dots,F_n\}
\end{equation}
For fault diagnosis syatem, the fault types in this work are represented by $F_i$ and the number of elements in the power set of $\chi$ is $2^{\mid\chi\mid}$. As an example, if there are three fault types $F_1$, $F_2$ and $F_3$, then the frame of discernment is $\{F_1,F_2,F_3\}$ and the power set is given as: $2^{\chi} = \{\phi, \{F_1\},\{F_2\}, \{F_3\}, \{F_1,F_2\}, \{F_1,F_3\}, \{F_2,F_3\},\newline\{F_1,F_2,F_3\}\}$. Here, $\{F_1\}$ shows that the $F_1$ fault has occurred, $\{F_1,F_2,F_3\}$ shows that either of the fault has occurred and $\phi$ represents no fault has occurred.

\textbf{Basic probability assignment} (BPA) represented by $m(.)$ is responsible for mapping each hypotheses (faults in this work) of frame of discernment that is $F_i$ to $m(F_i)$ $\in [0,1]$ with the conditions that:
\begin{equation}
m(\phi) = 0 \hspace{3mm}\text{and} \sum_{F_i\subseteq \chi} m(F_i) =1
\end{equation}
This reflects the support degree of $F_i$. As an example, if $m(F_1) = 0.6, m(F_2) = 0.3$ and $m(F_1,F_2)=0.1$, then if $\{F_1\}, \{F_2\}$ and $\{F_1,F_2\}$ happen, the support degrees are 0.6, 0.3 and 0.1 respectively.

$F_i$ is called the \textbf{focal element} (FE) if $m(F_i) > 0, F_i\subseteq \chi$. The FEs for fault types $F_1$, $F_2$ and $F_3$ are $\{F_1\},\{F_2\}, \{F_3\}, \{F_1,F_2\}, \{F_1,F_3\}, \{F_2,F_3\},\newline\{F_1,F_2,F_3\}$.

In Dempster-Shafer Theory, BPA can be generated from multiple sensors based on frame of discernment. These BPA can be combined together using Dempster's Combination Rule (DCR):
\begin{equation}
	m(F)=\begin{cases}\frac{\sum_{F_i\cap F_j=F} m_1(F_i)m_2(F_j)}{1-K}, & F\ne \phi \\
	= 0, & F = \phi
	
	\end{cases}
\end{equation}
where $K = \sum_{F_i\cap F_j=\phi} m_1(F_i)m_2(F_j)$ is the severity of the conflict; larger the value of K, greater is the conflict between the different evidences.
\subsection{Weighted Deng Entropy}
In Dempster-Shafer Evidence Theory, both BPA and frame of discernment are the reasons which attribute to uncertainties. The existing belief entropy like Deng entropy~\cite{DENG2016} and Dubois and Prade's weighted Hartley entropy~\cite{DUBOIS} consider only mass functions but not the frame of discernment. Weighted Deng entropy~\cite{TangY} was formulated to address these uncertainties. The scale of frame of discernment denoted as $\mid\chi\mid$ and the relative scale of the proposition or focal element with respect to the frame of discernment is $\frac{\mid F\mid}{\mid\chi\mid}$. Thus, the modified belief entropy can be given by:
\begin{equation}
	E_{wd}(m_i)=-\sum_{i} \frac{\mid F\mid m_i(F)}{\mid\chi\mid}log_2\frac{m_i(F)}{2^{\mid F \mid}-1}
\end{equation}
where, $F$ is a proposition or focal element of mass function $m$ and $\mid F\mid$ is the cardinality of proposition $F$.
\section{Proposed Method based on Weighted Deng Entropy}
\label{pm}
\begin{figure}[H]
	\centering
	\includegraphics[width=13cm,height=5cm]{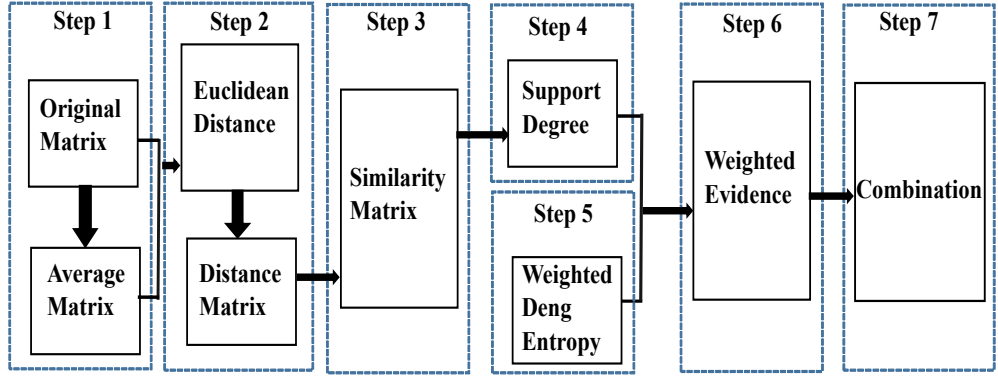} 
	\caption{Flowchart of the proposed method}
	\label{flowchart}
\end{figure}

The steps for the proposed improved Dempster Combination Rule (iDCR) are given as follows:
\begin{steps}
	\item If there is an evidence set $M=\{m_i|i=1,2,\dots,n\}$ in frame of discernment with focal elements $F=\{F_1,F_2,\dots,F_N\}$, then the average value $m_{avg}(F_j)$ can be defined as:
	\begin{equation}
	m_{avg}(F_j)=\frac{1}{n}\sum_{i=1}^{n}m_i(F_j), j=1,2,\dots,N
	\end{equation}
	where, $m_{avg}(F_j) \in [0,1]$ and $\sum_{j=1}^{N} m_{avg}(F_j) = 1$.
	\item Next, the Euclidean distance ($dist(m_i,m_{avg})$) between the original BPA and the average BPA is calculated by:
	\begin{equation}
	dist(m_i,m_{avg}) = \sqrt{\sum_{j=1}^{N}[m_i(F_j)-m_{avg}(F_j)]^2}, i=1,2,\dots,n
	\end{equation}
	The $n$ dimension distance vector can thus be represented as:
	\begin{equation*}
		D = 
		\begin{bmatrix}
			dist(m_1,m_{avg}) \\
			dist(m_2,m_{avg}) \\
			\dots\\
			dist(m_n,m_{avg})
		\end{bmatrix}
	\end{equation*}
	\item The similarity between two evidences can be derived from the previously calculated distance $dist(m_i,m_{avg})$ and is given by:
	\begin{equation}
	S(m_i,m_{avg})=1-dist(m_i,m_{avg})
	\end{equation}
	So, the $n$ dimension similarity vector can be represented as:
	\begin{equation*}
	Sim = 
	\begin{bmatrix}
	S(m_1,m_{avg}) \\
	S(m_2,m_{avg}) \\
	\dots\\
	S(m_n,m_{avg})
	\end{bmatrix}
	\end{equation*}
	\item Now, if the similarity between two evidences shows a higher mutual support degree, it indicates that the conflict is less. The support degree of evidence $m_i$ can thus be evaluated by:
	\begin{equation}
	sup(m_i)=\frac{S(m_i,m_{avg})}{\sum_{k=1}^{n}S(m_k,m_{avg})}
	\end{equation}
	\item Next, the weighted Deng entropy~\cite{TangY} is calculated as follows:
	\begin{equation}
	E_{wd}(m_i)=-\sum_{i} \frac{\mid F\mid m_i(F)}{\mid\chi\mid}log_2\frac{m_i(F)}{2^{\mid F \mid}-1}
	\end{equation}
	\item In this step, based on the support degree $S(m_i,m_{avg})$ and weighted Deng entropy $E_{wd}$, the reliability/credibility degree or the weight of $m_i$ is given by:
	\begin{equation}
	w_i =crd(m_i)= \frac{sup(m_i)*E_{wd}(m_i)}{\sum_{k=1}^{n}sup(m_k)*E_{wd}(m_k)}
	\end{equation}
	Here, $w_i$ is the weight of the $i^{th}$ evidence. Thus, the modified BPA can be given as:
	\begin{equation}
	m'(F_j) = \sum_{i=1}^{n}w_im_i(F_j),\hspace{8mm} j=1,2,\dots,N
	\end{equation}
	\item Finally, Dempster's combination rule is applied to combine the modified BPA.
	\begin{equation}
	m(F)=\begin{cases}\frac{\sum_{\cap F_j=F} \prod_{i=1}^{n}m_i'(F_j)}{1-\sum_{\cap F_j=\phi} \prod_{i=1}^{n}m_i'(F_j)}, &F\ne \phi, \forall F_j \subseteq \chi\\
	0, & F = \phi
	
	\end{cases}
	\end{equation}
	Here, m(F) is the final fusion result.
\end{steps}
A diagrammatic representation of the steps is given in Fig.~\ref{flowchart}.

\subsection{Examples}
\begin{Example}
	\label{ex1}
	A frame of discernment of a fault diagnosis system is $\chi = \{F_1, F_2, F_3\}$. The BPAs ($m_1$ and $m_2$) from two different sensors are given in Table~\ref{tab_1}.
	\begin{table}[H]
		\setlength{\tabcolsep}{5.5pt}
	       	\centering
		\vspace{-0.9em}    
		\caption{BPA from two sensors}
		\label{tab_1}
		\resizebox{0.4\textwidth}{!}{
			\begin{tabular}{l l l l}
			
				\hline
				\hline
				Evidence & $F_1$ & $F_2$ & $F_3$ \\
				\cline{1-4}
				$m_1$ & 0.99 & 0.01 & 0\\
				$m_2$ & 0 & 0.01 & 0.99\\
				\hline
				\hline
				
			\end{tabular}
		}
	\end{table}
\end{Example}
From Table~\ref{tab_1}, it can be seen that the support degree of $F_1$ from the first sensor is 0.99 and that from the second sensor is 0. For $F_3$ on the other hand, the reverse is true. The support degree is thus clearly conflicting. Again, for $F_2$ the support degree for both the sensors is 0.01 which indicates that this fault is not very likely to occur. But the fusion result from the classical DCR is $m(F_1) = 0$, $m(F_2) = 1$, $m(F_3) = 0$ and $m(\chi) = 0$. Thus, the support degree according to classical DCR is 1 indicating that fault $F_2$ will definitely happen which is of course not supported by the actual condition. So, it can be easily concluded that if there is a conflict between evidences, classical DCR shows paradoxical result. 

Now, that the flaw in the traditional method has been established, the calculation for Example~\ref{ex1} for the proposed method is given in details as follows:
\begin{steps}
	\item First, the average values are calculated:
	$$m_{avg}(F_1)=\frac{1}{2}\sum_{i=1}^{2}m_i(F_1)=\frac{1}{2}(0.99+0)=0.495$$
	$$m_{avg}(F_2)=\frac{1}{2}\sum_{i=1}^{2}m_i(F_2)=\frac{1}{2}(0.01+0.01)=0.01$$
	$$m_{avg}(F_3)=\frac{1}{2}\sum_{i=1}^{2}m_i(F_3)=\frac{1}{2}(0+0.99)=0.495$$
	\item In this step, the Euclidean distances are calculated:
	$$	dist(m_1,m_{avg}) = \sqrt{\sum_{j=1}^{3}[m_1(F_j)-m_{avg}(F_j)]^2}$$
	$$=\sqrt{0.495^2+0^2+0.495^2} = 0.7$$
	$$	dist(m_2,m_{avg}) = \sqrt{\sum_{j=1}^{3}[m_2(F_j)-m_{avg}(F_j)]^2}$$
	$$=\sqrt{0.495^2+0^2+0.495^2} = 0.7$$
	\item Next, the similarity degrees between the evidences are calculated:
	$$S(m_1,m_{avg})=1-dist(m_1,m_{avg})=1-0.7=0.3$$
	$$S(m_2,m_{avg})=1-dist(m_2,m_{avg})=1-0.7=0.3$$
	\item From the similarity degrees, support degrees are evaluated:
	$$sup(m_1)=\frac{S(m_1,m_{avg})}{\sum_{k=1}^{2}S(m_k,m_{avg})}=0.5$$
	$$sup(m_2)=\frac{S(m_2,m_{avg})}{\sum_{k=1}^{2}S(m_k,m_{avg})}=0.5$$
	\item The weighted Deng entropy is calculated next:
	$$E_{wd}(m_1)=-\sum_{i=1} \frac{\mid F\mid m_1(F)}{\mid\chi\mid}log_2\frac{m_1(F)}{2^{\mid F \mid}-1}=0.0202$$
	$$E_{wd}(m_2)=-\sum_{i=2} \frac{\mid F\mid m_2(F)}{\mid\chi\mid}log_2\frac{m_2(F)}{2^{\mid F \mid}-1}=0.0202$$
	\item The credibility degree or the weight and the modified BPA are deduced in this step:
	$$w_1 =crd(m_1)= \frac{sup(m_1)*E_{wd}(m_1)}{\sum_{k=1}^{n}sup(m_k)*E_{wd}(m_k)}=0.5$$
	$$w_2 =crd(m_2)= \frac{sup(m_2)*E_{wd}(m_2)}{\sum_{k=1}^{n}sup(m_k)*E_{wd}(m_k)}=0.5$$
	$$m'(F_1) = \sum_{i=1}^{2}w_im_i(F_1)=0.495$$
	$$m'(F_2) = \sum_{i=1}^{2}w_im_i(F_2)=0.01$$
	$$m'(F_3) = \sum_{i=1}^{2}w_im_i(F_3)=0.495$$
	\item Finally, DCR is applied to calculate the fusion result: 
	$m(F_1) = 0.4999$, $m(F_2) = 0.0002$ and $m(F_3) = 0.4999$. This result is a clear indication that the faults $F_1$ and $F_3$ are much more likely to happen than fault $F_2$, thus supporting the real condition, making the proposed method more feasible.
 \end{steps}
 \begin{Example}
 	\label{ex2}
 	A frame of discernment $\chi$ has 7 elements, of which $F_1, F_2$ and $F_3$ are three different faults. The BPAs ($m_1$, $m_2$ and $m_3$) are given in Table~\ref{tab_2}.
 	\begin{table}[H]
 		\setlength{\tabcolsep}{5.5pt}
 		\centering
 		\vspace{-0.9em}    
 		\caption{BPA from three sensors}
 		\label{tab_2}
 		\resizebox{0.7\textwidth}{!}{
 			\begin{tabular}{l l l l l l l l}
 				
 				\hline
 				\hline
 				Evidence & $\{F_1\}$ & $\{F_2\}$ & $\{F_3\}$ & $\{F_1,F_2\}$ & $\{F_2,F_3\}$ & $\{F_1,F_3\}$ & $\{F_1,F_2,F_3\}$\\
 				\cline{1-8}
 				$m_1$ & 0.70 & 0.05 & 0.05 &0.05& 0.05& 0.05 &0.05\\
 				$m_2$ & 0.05 & 0.70 & 0.05 &0.05& 0.05& 0.05 &0.05\\
 				$m_3$ & 0.75  &0.05& 0& 0.05& 0.05& 0.05& 0.05\\
 					\hline
 					\hline
 				\end{tabular}
 		}
 	\end{table}
  \end{Example}	
 	\begin{table}[H]
 		\setlength{\tabcolsep}{5.5pt}
 		\tiny
 		\centering
 		\vspace{-0.9em}    
 		\caption{Fusion Results}
 		\label{tab_3}
 		\resizebox{0.6\textwidth}{!}{
 			\begin{tabular}{l l l l}
 				
 					\hline
 					\hline
 				
 				Rules & Evidence & $m_{1,2}$ & $m_{1,2,3}$\\
 				\cline{1-4}
 				 & $m(F_1)$ & 0.4579 & 0.8381\\
 				Wang et. al~\cite{ZWang}& $m(F_2)$ & 0.4579 & 0.1431\\
				& $m(F_3)$ & 0.0399 & 0.0114\\
 				& $m(F_1,F_2)$ & 0.0133 & 0.0023\\
 				& $m(F_1,F_3)$ & 0.0133 & 0.0023\\
				& $m(F_2,F_3)$ & 0.0133 & 0.0023\\
 				& $m(F_1,F_2,F_3)$ & 0.0044 & 0.0005\\
 				\cline{1-4}
 				& $m(F_1)$ & 0.4787 & 0.9542\\
 				iDCR & $m(F_2)$ & 0.4787 & 0.0430\\
 				& $m(F_3)$ & 0.0085 & 0.0001\\
 				& $m(F_1,F_2)$ & 0.0085 & 0.0006\\
 				& $m(F_1,F_3)$ & 0.0085 & 0.0006\\
 				& $m(F_2,F_3)$ & 0.0085 & 0.0006\\
 				& $m(F_1,F_2,F_3)$ & 0.0085 & 0.0006\\
 				\hline
 				\hline
 			\end{tabular}
 		}
 	\end{table}

\begin{figure}[H]
	\centering
	\begin{subfigure}{0.5\textwidth}
		\includegraphics[width=6.3cm,height=6.3cm]{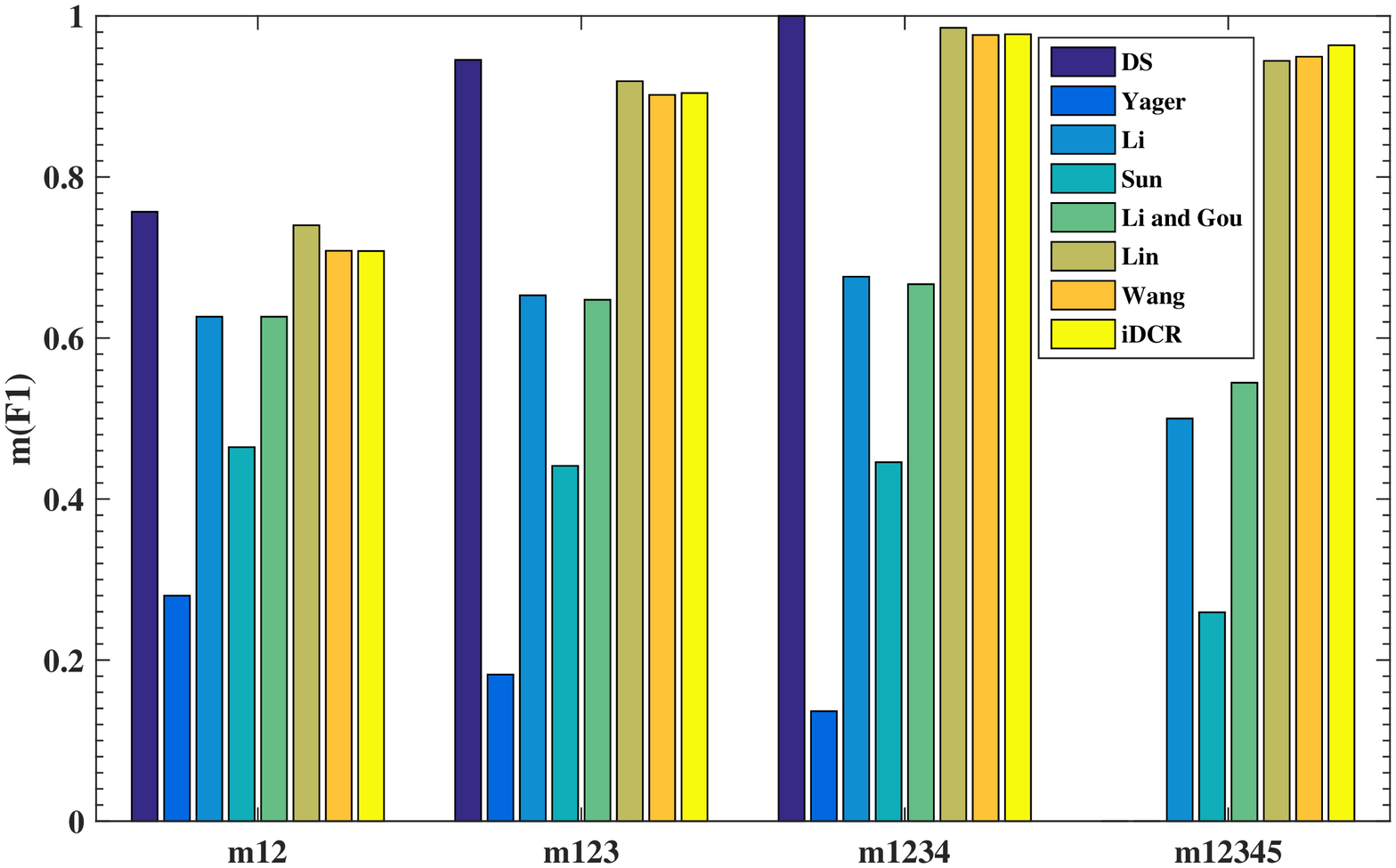} 
		\caption{}
		\label{fig1a}
	\end{subfigure}\begin{subfigure}{0.5\textwidth}
		\includegraphics[width=6.4cm,height=6.3cm]{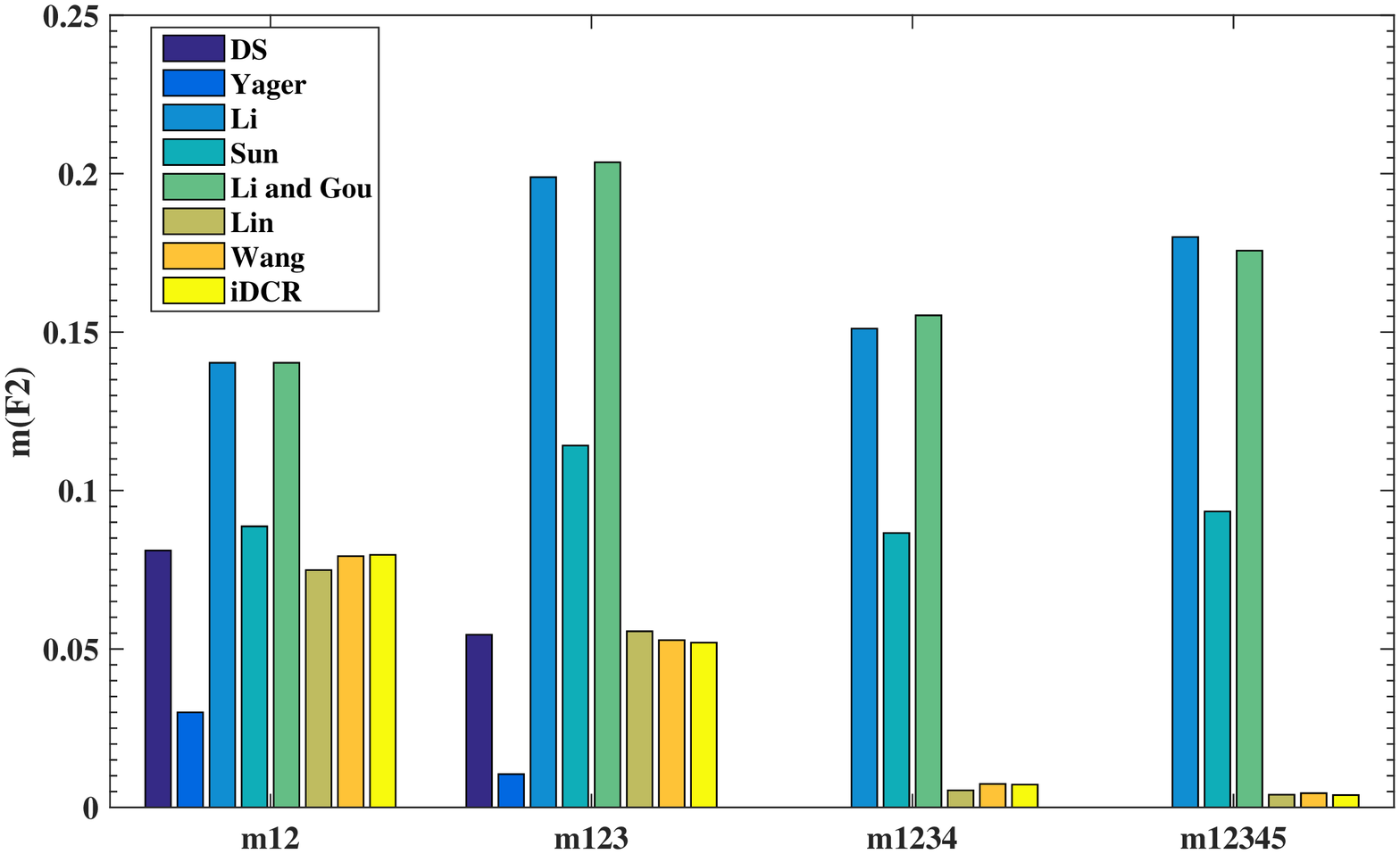}
		\caption{}
		\label{fig1b}
	\end{subfigure}
	\begin{subfigure}{0.5\textwidth}
	
		\includegraphics[width=6.4cm,height=6.4cm]{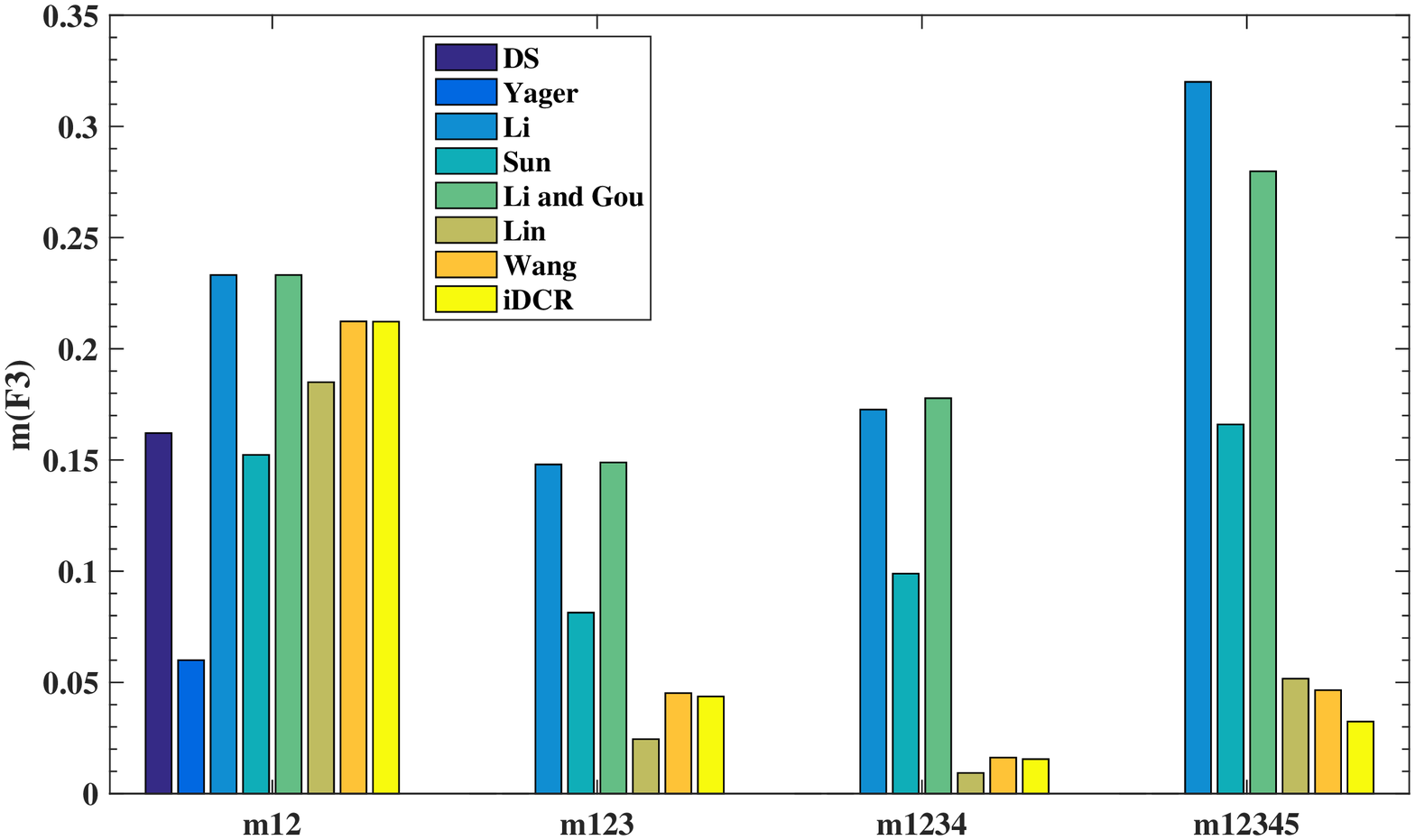}
		\caption{}
		\label{fig1c}
	\end{subfigure}
	\caption{(a) Combination results for $F_1$, (b) Combination results for $F_2$, (c) Combination results for $F_3$}
	\label{fig:fig1}
\end{figure}

It can be deduced from Table~\ref{tab_3} that, though there is huge conflict in the data, for both~\cite{ZWang} and the proposed method the fusion result of $F_1$ is the highest when all the three sensors are combined. Though both the methods give the correct judgement, the support degree of $F_1$ for the proposed method is higher, proving its superiority over~\cite{ZWang}.
\begin{Example}
\label{ex3}
A frame of discernment of a fault diagnosis system is $\chi= \{F_1,F_2,F_3\}$ and the BPA is given as follows:
\begin{table}[H]
	\setlength{\tabcolsep}{5.5pt}
	\tiny
	\centering
	\vspace{-0.9em}    
	\caption{BPA from three sensors}
	\label{tab_4}
	\resizebox{0.4\textwidth}{!}{
		\begin{tabular}{l l l l l}
			\hline
			\hline
			
			Evidence & $F_1$ & $F_2$ & $F_3$ & $\chi$\\
			\cline{1-5}
			$m_1$ & 0.70 & 0.15 & 0.15 &0\\
			$m_2$ & 0.40 & 0.20 & 0.40 &0\\
			$m_3$ & 0.65  &0.35& 0& 0\\
			$m_4$ & 0.75  &0& 0.25& 0\\
			$m_5$ & 0  &0.20& 0.80& 0\\
			\hline
			\hline

		\end{tabular}
	}
\end{table}
\end{Example}	
\begin{table}[H]
	\setlength{\tabcolsep}{5.5pt}
	\tiny
	\centering
	\vspace{-0.9em}    
	\caption{Fusion Results}
	\label{tab_5}
	\resizebox{0.8\textwidth}{!}{
		\begin{tabular}{l l l l l l}
			
			\hline
			\hline
			
			Rules & Evidence & $m_{1,2}$ & $m_{1,2,3}$ & $m_{1,2,3,4}$ & $m_{1,2,3,4,5}$\\
			\cline{1-6}
			& $m(F_1)$ & 0.7568 & 0.9455 & 1 & NaN\\
			DS~\cite{LIUChen} & $m(F_2)$ & 0.0811 & 0.0545 & 0 & NaN\\
			& $m(F_3)$ & 0.1621 & 0 & 0 & NaN\\
			& $m(\chi)$ & 0 & 0 & 0 & NaN\\
			\cline{1-6}
			& $m(F_1)$ & 0.2800 & 0.1820 & 0.1365 & 0\\
			Yager~\cite{yager} & $m(F_2)$ & 0.0300 & 0.0105 & 0 & 0\\
			& $m(F_3)$ & 0.0600 & 0 & 0 & 0\\
			& $m(\chi)$ & 0.6300 & 0.8075 & 0.8635 & 1\\
			\cline{1-6}
			& $m(F_1)$ & 0.6265 & 0.6531 & 0.6762 & 0.5000\\
			Li \textit{et al}.~\cite{li2001efficient} & $m(F_2)$ & 0.1403 & 0.1989 & 0.1511 & 0.1800\\
			& $m(F_3)$ & 0.2332 & 0.1480 & 0.1727 & 0.3200\\
			& $m(\chi)$ & 0 & 0 & 0 & 0\\
			\cline{1-6}
			& $m(F_1)$ & 0.4645 & 0.4412 & 0.4457 & 0.2594\\
			Sun \textit{et al}.~\cite{Sun} & $m(F_2)$ & 0.0887 & 0.1142 & 0.0866 & 0.0934\\
			& $m(F_3)$ & 0.1523 & 0.0814 & 0.0989 & 0.1660\\
			& $m(\Theta)$ & 0.2945 & 0.3632 & 0.3688 & 0.4812\\
			\cline{1-6}
			& $m(F_1)$ & 0.6265 & 0.6475 & 0.6669 & 0.5445\\
			Li and Gou~\cite{WLLi} & $m(F_2)$ & 0.1403 & 0.2036 & 0.1553 & 0.1757\\
			& $m(F_3)$ & 0.2332 & 0.1489 & 0.1778 & 0.2798\\
			& $m(\chi)$ & 0 & 0 & 0 & 0\\
			\cline{1-6}
			& $m(F_1)$ & 0.7401 & 0.9190 & 0.9854 & 0.9443\\
			Lin \textit{et al}.~\cite{YLin} & $m(F_2)$ & 0.0749 & 0.0556 & 0.0054 & 0.0040\\
			& $m(F_3)$ & 0.1850 & 0.0245 & 0.0093 & 0.0517\\
			& $m(\chi)$ & 0 & 0 & 0 & 0\\
			\cline{1-6}
			& $m(F_1)$ & 0.7084 & 0.9020 & 0.9764 & 0.9495\\
			Wang \textit{et al}.~\cite{ZWang} & $m(F_2)$ & 0.0793 & 0.0528 & 0.0074 & 0.0045\\
			& $m(F_3)$ & 0.2123 & 0.0452 & 0.0162 & 0.0465\\
			& $m(\chi)$ & 0 & 0 & 0 & 0\\
			\cline{1-6}
			& $m(F_1)$ & 0.7081 & 0.9043 & 0.9773 & 0.9637\\
			iDCR & $m(F_2)$ & 0.0797 & 0.0520 & 0.0072 & 0.0039\\
			& $m(F_3)$ & 0.2122 & 0.0437 & 0.0155 & 0.0324\\
			& $m(\chi)$ & 0 & 0 & 0 & 0\\
			\hline
			\hline
		\end{tabular}
	}
\end{table}
From Table~\ref{tab_4}, it can be seen that sensors 1,2,3 and 4 support the fault $F_1$ whereas fault $F_3$ is supported by the sensor 5. Additionally, according to the data from sensor 5, fault $F_1$ does not happen. This shows that there is high conflict in the sensor data. Also, as no other sensor supports fault $F_3$ other than sensor 5, when fusion rule is applied, the values of $m(F_1)$ should be higher than both $m(F_2)$ and $m(F_3)$. From Table~\ref{tab_5} and Fig.~\ref{fig:fig1} it is evident that as the number of evidences increases, $m(F_1)$ also increases but $m(F_2)$ and $m(F_3)$ decrease. Though traditional DCR follows this trend, but it becomes useless when there is high conflict, that is, when the fifth evidence is introduced. Yager's method is unable to resolve the high conflict situation as well as it removes the normalisation process in DCR and also assigns the conflict to an unknown domain. To allocate the conflict in evidence, Li \textit{et al}. has set up a model whereas Sun \textit{et al}. has considered only partial evidences leading to strong uncertainty in results. By modifying the model, Li and Gou distributed the conflict resulting conservative fusion result. So, from the results it can be deduced that most of the methods adhere to the view that the second evidence is the most credible while the reverse is true for the fifth evidence. The proposed method (iDCR) performs better while dealing with information with high conflict; fusion result of fault $F_1$ has a slight decrease from 0.9773 to 0.9637, as opposed to Lin \textit{et al}. and Wang \textit{et al}. where the decrease is from 0.9854 to 0.9443 and 0.9764 to 0.9495 respectively.
 \section{Application of Fault Diagnosis Method}
 \label{app}
The fault diagnosis architecture of a multi sensor fusion system used in this work is described below:

\subsection{Fault Diagnosis Architecture} 
Data from multiple sensors (which are distributed in the system) have been used in this work to improve the accuracy of the system in terms of fault diagnosis. The sensors can generate their own evidences for different faults when they are working. Later, DCR is used to combine all these evidences as have been discussed earlier. Finally, decision rules are applied to make the final decision about the system. 

The system is divided into five levels to achieve its goals, they being:
\begin{enumerate}
	\item Data Level: In this level, data is collected from the different sensors and the frame of discernment is built.
	\item Feature Level: Different fault features are extracted in this level to monitor the system status.
	\item Evidence Level: Depending on the extracted features, BPA of different sensors are generated at this level.
	\item Fusion Level: As the name suggests, different BPAs are combined in this level.
	\item Decision Level: Finally, in this level feasible decision rule is applied to derive the final conclusion.
\end{enumerate}	
\subsection{Testing}
A rotating 	machinery~\cite{YLin} is used in this work to verify the performance of the system. The faults in the  system have been categorised into four types: $F_1$ = ``Imbalance", $F_2$ = ``Shaft crack", $F_3$ = ``Misalignment" and $F_4$ = ``Bearing loose". Thus, the frame of discernment is $\chi = \{F_1, F_2, F_3, F_4\}$.

In the \textbf{data level} five sensors have been used to monitor the status of the system. Next, in \textbf{feature level} the fault features considered are: $E_1$ = ``Power spectrum entropy", $E_2$ = ``Wavelet energy spectrum entropy", $E_3$ = ``Wavelet space spectral entropy" and $E_4$ = ``Singular spectrum entropy". The reference fault feature vector is $\rho(F_i) = \{E_{i1}, E_{i2}, E_{i3}, E_{i4}\} (i=1,2,3,4)$ as procured from the training data. In \textbf{evidence level}, the distance between the reference and the measurement fault feature vector for five sensors is:
\begin{equation}
d_{ji} = d_j(F_i) = [\sum_{j=1}^{5}\mid\omega_j(F)-\rho(F_i)\mid]^{1/2}
\end{equation}
where, $\omega_j(F) = \{E_1',E_2',E_3,E_4'\} (j=1,2,\dots,5)$. 

As, the distance function and the similarity degree are reciprocal to each other, thus to transform the distance function to similarity degree, an inverse function has been used. Then the BPA of the $j^{th}$ sensor can be calculated by normalising the similarity degree. Next in the \textbf{fusion level}, the proposed method is used to fuse the different pieces of evidence. Finally, in the \textbf{decision level} if $\forall F_1,F_2 \subset \chi$ satisfy:
\begin{equation}
\begin{cases} m(F_1)=max\{m(F_i),F_i \subset \chi\}\\
m(F_2)=max\{m(F_i),F_i \subset \chi, F_i\ne F_1\}
\end{cases}
\end{equation}
$\xi_1$ and $\xi_2$ are the two thresholds of decision and $F_1$ is the final decision if:
\begin{equation}
\begin{cases} m(F_1) - m(F_2) > \xi_1\\
m(\chi)< \xi_2\\
m(F_1) > m(\chi)
\end{cases}
\end{equation}
\begin{Example}
	Reference fault features of four mechanical faults are given in Table~\ref{tab_6} according to~\cite{YLin}. \begin{table}[H]
		\setlength{\tabcolsep}{5.5pt}
		\tiny
		\centering
		\vspace{-0.9em}    
		\caption{Reference Fault Feature Vector}
		\label{tab_6}
		\resizebox{0.6\textwidth}{!}{
			\begin{tabular}{l l l l l}
				\hline
				\hline
				
			Fault Types & $E_1$ & $E_2$ & $E_3$ & $E_4$\\
				\cline{1-5}
				$F_1$ & 43.5828 & 30.8859 & 10.6806 & 53.7373\\
				$F_2$ & 74.3605 & 72.1393 & 17.8107 & 74.1857\\
				$F_3$ & 63.9286  & 58.6064 & 21.7660 & 67.5529\\
				$F_4$ & 49.8858  &46.8183& 14.998& 52.6699\\
				\hline
				\hline

			\end{tabular}
		}
	\end{table}
	\begin{table}[H]
		\setlength{\tabcolsep}{5.5pt}
		\tiny
		\centering
		\vspace{-0.9em}    
		\caption{Fault Feature of Five Sensors}
		\label{tab_7}
		\resizebox{0.6\textwidth}{!}{
			\begin{tabular}{l l l l l}
				\hline
				\hline
				
				Sensors & $E_1$ & $E_2$ & $E_3$ & $E_4$\\
				\cline{1-5}
				Sensor 1 & 66.2913 & 57.3129 & 22.8701 & 65.0923\\
				Sensor 2 & 62.3361 & 55.3681 & 22.8297 & 66.1382\\
				Sensor 3 & 73.4274  & 69.8329 & 16.5621 & 72.5824\\
				Sensor 4 & 65.8638  & 61.5325 & 24.2016 & 69.2899\\
				Sensor 5 & 51.4154  &48.3248& 15.4123 & 50.3624\\
				\hline
				\hline

			\end{tabular}
		}
	\end{table}
	\begin{table}[H]
		\setlength{\tabcolsep}{5.5pt}
		\tiny
		\centering
		\vspace{-0.9em}    
		\caption{BPA of five sensors}
		\label{tab_8}
		\resizebox{0.6\textwidth}{!}{
			\begin{tabular}{l l l l l}
				\hline
				\hline
				
				Evidence & $F_1$ & $F_2$ & $F_3$ & $F_4$\\
				\cline{1-5}
				$m_1$ & 0.1469 & 0.2057 & \textit{\textbf{0.4660}} & 0.1813\\
				$m_2$ & 0.1521 & 0.1935 & \textit{\textbf{0.4631}} & 0.1914\\
				$m_3$ & 0.1278  & \textit{\textbf{0.5008}} & 0.2221 & 0.1493\\
				$m_4$ & 0.1459  & 0.2396 & \textit{\textbf{0.4395}} & 0.1750\\
				$m_5$ & 0.2068  &0.1399& 0.1755 & \textit{\textbf{0.4777}}\\
				\hline
				\hline

			\end{tabular}
		}
	\end{table}
\end{Example}
\begin{table}[H]
	\setlength{\tabcolsep}{5.5pt}
	\tiny
	\centering
	\vspace{-0.9em}    
	\caption{Fusion Results}
	\label{tab_9}
	\resizebox{0.8\textwidth}{!}{
		\begin{tabular}{l l l l l l}
			
			\hline
			\hline
			
			Rules & Results & $F_1$ & $F_2$ & $F_3$ & $F_4$\\
			\cline{1-6}
			DS & & 0.0714 & 0.1273 & 0.6902 & 0.1110\\
			Lin \textit{et al}.~\cite{YLin} & $m_{1,2}$ & 0.0715 & 0.1274 & 0.6903 & 0.1111\\
			Wang \textit{et al}.~\cite{ZWang} & & 0.0715 & 0.1274 & 0.6900 & 0.1110\\
			iDCR & & 0.0715 & 0.1274 &\textbf{0.6901} & 0.1111\\
			\cline{1-6}
			DS & & 0.0376 & 0.2626 & 0.6315 & 0.0683\\
			Lin \textit{et al}.~\cite{YLin} & $m_{1,2,3}$ & 0.0315 & 0.2675 & 0.6431 & 0.0579\\
			Wang \textit{et al}.~\cite{ZWang} & & 0.0314 & 0.2594 & 0.6490 & 0.0578\\
			iDCR & & 0.0315 & 0.2540 & \textbf{0.6565} & 0.0585\\
			\cline{1-6}
			DS & & 0.0153 & 0.1758 & 0.7755 & 0.0334\\
			Lin \textit{et al}.~\cite{YLin} & $m_{1,2,3,4}$ & 0.0125 & 0.1692 & 0.7906 & 0.0276\\
			Wang \textit{et al}.~\cite{ZWang} & & 0.0126 & 0.1643 & 0.8026 & 0.0278\\
			iDCR & & 0.0124 & 0.1571 & \textbf{0.8029} & 0.0275\\
			\cline{1-6}
			DS & & 0.0176 & 0.1368 & 0.7570 & 0.0886\\
			Lin \textit{et al}.~\cite{YLin} & $m_{1,2,3,4,5}$ & 0.0109 & 0.1258 & 0.7874 & 0.0759\\
			Wang \textit{et al}.~\cite{ZWang} & & 0.0108 & 0.1204 & 0.7941 & 0.0747\\
			iDCR & & 0.0103 & 0.1148 & \textbf{0.8011} & 0.0692\\
			\hline
			\hline
		\end{tabular}
	}
\end{table}
 Table~\ref{tab_7} provides the fault feature of the five sensors. Table~\ref{tab_8} gives the BPA of the five sensors at the evidence level from which it can be seen that the fault $F_3$ should happen. But the conflict information says otherwise. Like, the informations from sensors 1,2 and 4 show that the fault is $F_3$, whereas sensor 3 and sensor 5 indicate that the fault is $F_2$ and $F_4$ respectively.
 \begin{figure}[H]
 	\centering
 	\includegraphics[width=8cm,height=8cm]{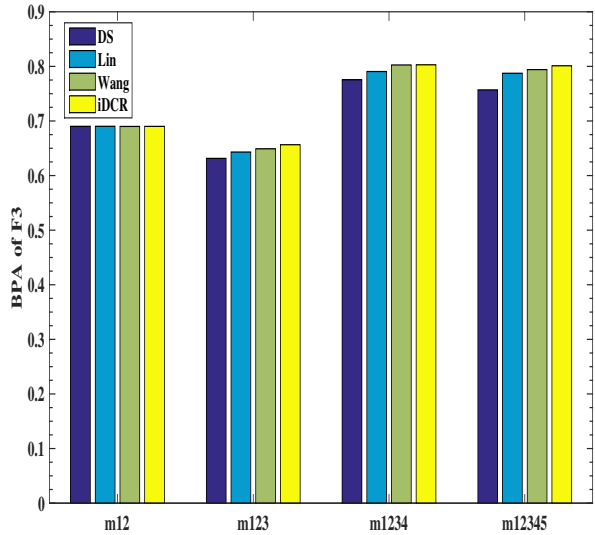} 
 	\caption{Fusion results of $F_3$}
 	\label{BPA}
 \end{figure}
 
 In Table~\ref{tab_9}, the calculations using the traditional DCR, some existing methods~\cite{YLin,ZWang} and the proposed method have been shown. Now, considering $\xi_1 = \xi_2 = 0.1$, from Table~\ref{tab_9}, it can be seen that all the methods confirm fault $F_3$. Though, when combining sensors 1, 2 and 3, there is a slight decrease in the support degree because of the conflicting information from sensor 3, fault $F_3$ prevails. The support degree again increases for fault $F_3$ when sensor 4 is fused. Finally, when all the five sensors are combined, fault $F_3$ is prevails. So, it can be concluded that the fault diagnosis system can be used to make the correct decision. Moreover, the superiority of the proposed method (iDCR) is evident from Table~\ref{tab_9} and Fig.~\ref{BPA}, as they show that iDCR has a higher support degree for fault $F_3$ as compared to classical DCR and existing methods.
 
 \section{Conclusion}
 \label{con}
 Multiple sensors are used for monitoring many systems where data from a single sensor is not enough or adequate. But the data collected from these multiple sensors may show conflicting results or may be uncertain, fuzzy or even incomplete, resulting in misleading conclusions when data from the multiple sensors are fused. In this work, a multisensor fault diagnosis method has been put forward considering the evidence theory. The proposed method (iDCR) considers weighted Deng entropy to improve upon the traditional Dempster's combination rule. The superiority of the proposed method has been shown through numerical examples and simulations taking different situations into account. As a future work, this work can be extended for more dynamic and complex environments and also the focus will be on further improving the fusion result. 
\bibliographystyle{plain}

\bibliography{mybibfile}

\end{document}